\documentclass[letterpaper,9pt]{IEEEtran}
\usepackage{cite}
\usepackage{amsmath,amssymb,amsfonts}
\usepackage{algorithmic}
\usepackage{graphicx}
\usepackage{textcomp}
\usepackage{xcolor}
\usepackage{mathrsfs}

\usepackage{amsfonts}
\usepackage{cite}
\usepackage{booktabs}

\usepackage{indentfirst}

\usepackage{flushend}

\usepackage{multirow}
\usepackage{float}
\usepackage{algorithm}
\usepackage{algorithmic}
\usepackage{nomencl}

\usepackage{etoolbox}

\usepackage{amssymb,pifont}

\usepackage{color}
\newcommand{\cmark}{\ding{51}} 
\newcommand{\xmark}{\ding{55}} 

\allowdisplaybreaks

\definecolor{rankone}{HTML}{DDF1D4}
\definecolor{ranktwo}{HTML}{FFFAD4}
\definecolor{rankthree}{HTML}{C2DEEB}

\definecolor{Gray}{gray}{0.95}
\definecolor{DarkRed}{rgb}{0.7, 0.0, 0.0} 
\usepackage{xcolor}
\usepackage{colortbl} 
\usepackage{arydshln}
\newcommand{\droptxt}[1]{\textcolor{DarkRed}{\fontsize{5pt}{5pt}\selectfont$\downarrow$#1\%}}

\newlength{\annNumW}
\newlength{\annPostW}

\newcommand{\cellN}[1]{%
	\makebox[\annNumW][r]{#1}\makebox[\annPostW][l]{}
}

\newcommand{\cellD}[2]{%
	\makebox[\annNumW][r]{#1}\makebox[\annPostW][l]{\hspace{1pt}\droptxt{#2}}
}

\newcommand{\rkbox}[2]{%
	\begingroup
	\setlength{\fboxsep}{1pt}
	\colorbox{#1}{\rule[-0.2ex]{0pt}{2ex}$#2$}%
	\endgroup
}

\newcommand{\rkone}[1]{\rkbox{rankone}{#1}}    
\newcommand{\rktwo}[1]{\rkbox{ranktwo}{#1}}    
\newcommand{\rkthree}[1]{\rkbox{rankthree}{#1}}

\begin{document}

\begin{@twocolumnfalse}
	
	\title{
		\Large \textbf{Efficient Sparse-to-Dense Visual Localization via Compact Gaussian Scene Representation and Accelerated Dense Pose Estimation}
	}
	
	\author{Zizhuo Li, Songchu Deng, Linfeng Tang, and Jiayi Ma
		\thanks{Corresponding author: Jiayi Ma.}
		\thanks{Zizhuo Li and Songchu Deng contributed equally to this work.}
		\thanks{This work was supported by the National Natural Science Foundation of China under Grant 62276192.}
		\thanks{Zizhuo Li and Songchu Deng are with the Electronic Information School, Wuhan University, Wuhan 430072, China. Linfeng Tang is with the School of Robotics, Wuhan University, Wuhan 430072, China. Jiayi Ma is with the Electronic Information School and the School of Robotics, Wuhan University, Wuhan 430072, China (e-mail: jyma2010@gmail.com).}
	}
	\maketitle
\end{@twocolumnfalse}

%

\noindent Dear Editor,\\

This letter presents LiteLoc, a novel and efficient localizer built on 3D Gaussian Splatting (3DGS). Previous state-of-the-art (SoTA) sparse-to-dense localizer, STDLoc, has shown remarkable localization capability but suffers from severe storage redundancy and computational latency. By revisiting its design decisions, we derive two simple yet highly effective improvements that cumulatively make LiteLoc much more efficient in both memory and computation, while also being easier to train. One key observation is that the color field, inherited directly from Feature 3DGS, is functionally useless for localization. Yet, its reconstruction of high-frequency photometric details necessitates excessive Gaussian primitives, resulting in a tightly coupled color–feature representation with significant memory overhead and sub-optimal feature field optimization. To resolve this, we propose a color-free decoupled feature field that constructs a compact Gaussian scene representation by retaining only task-essential feature attributes, thereby eliminating $\approx 94\%$ of redundant storage with no loss of localization-relevant information. We further find that the primary computational bottleneck lies in the dense Perspective-n-Point (PnP) solver where most matches contribute saturated geometric constraints with diminishing accuracy gains. Accordingly, we propose a condensing strategy that distills any dense matches into a subset of 5\% \emph{representative matches}, enabling a nearly $19\times$ speedup in robust estimation with negligible performance drop. Extensive experiments show that LiteLoc surpasses STDLoc in multiple scenes with considerable efficiency benefits, opening up exciting prospects for latency-sensitive visual localization.

{\bf Introduction:} Visual localization, estimating the 6-DoF camera pose of a query image in a pre-built scene map, is essential for augmented reality, autonomous driving, and robotics~\cite{chen2023deep}. An appropriate scene representation is pivotal to the accuracy, robustness, and scalability of visual localization. Traditional feature matching (FM)~\cite{lu2024feature}-based methods~\cite{sattler2016efficient,sarlin2019coarse} rely on the sparse 3D point cloud reconstructed via Structure-from-Motion (SfM)~\cite{schonberger2016structure}, where each 3D point is associated with local descriptors~\cite{detone2018superpoint}. During localization, keypoints are extracted from the query image and matched either with reference images or directly with the 3D point cloud, followed by pose estimation using a PnP solver based on established 2D-3D matches. While effective in texture-rich environments, such methods degrade markedly in texture-poor scenarios due to insufficient reliable matches. Moreover, this paradigm is inherently storage-intensive, as retaining 3D points, descriptors, and co-visibility relationships incurs a substantial memory burden, especially for large-scale scenes. Alternatives encode scene information into deep neural networks, regressing either absolute poses (APR)~\cite{kendall2015posenet} or dense scene coordinates (SCR)~\cite{brachmann2021visual,brachmann2023accelerated,tang2023neumap}. The APR paradigm often suffers from limited accuracy and generalization to unseen views. The SCR paradigm excels in small-scale indoor weak-texture environments by predicting dense matches, yet falls short of handling large-scale outdoor scenes due to its fixed network capacity. Recently, implicit scene representations like NeRF~\cite{chen2024neural,moreau2023crossfire,zhao2024pnerfloc,zhou2024nerfect} and explicit ones like 3DGS~\cite{huang2025sparse,khatib2025gsvisloc} have emerged as powerful tools for visual localization, where their high-fidelity novel view synthesis capability is leveraged for data augmentation and iterative pose refinement. Building on this line of research, most recently, STDLoc~\cite{huang2025sparse} stands out by introducing a full sparse-to-dense pipeline based on Feature 3DGS~\cite{zhou2024feature}. Specifically, in the mapping stage, STDLoc constructs a feature-distilled 3D Gaussian scene representation by jointly optimizing geometric parameters, color attributes, and high-dimensional features for each Gaussian primitive. At inference time, given a query image, sparse 2D–3D correspondences are established by directly matching detected 2D keypoints with sampled 3D landmark Gaussians, for coarse pose initialization. Then, the localization accuracy is improved by aligning the query image's dense feature map to the trained feature field through dense matching. By doing so, STDLoc is robust to illumination changes and viewpoint variations. 

\begin{figure}[t]
	\centering
	\includegraphics[width=1\linewidth]{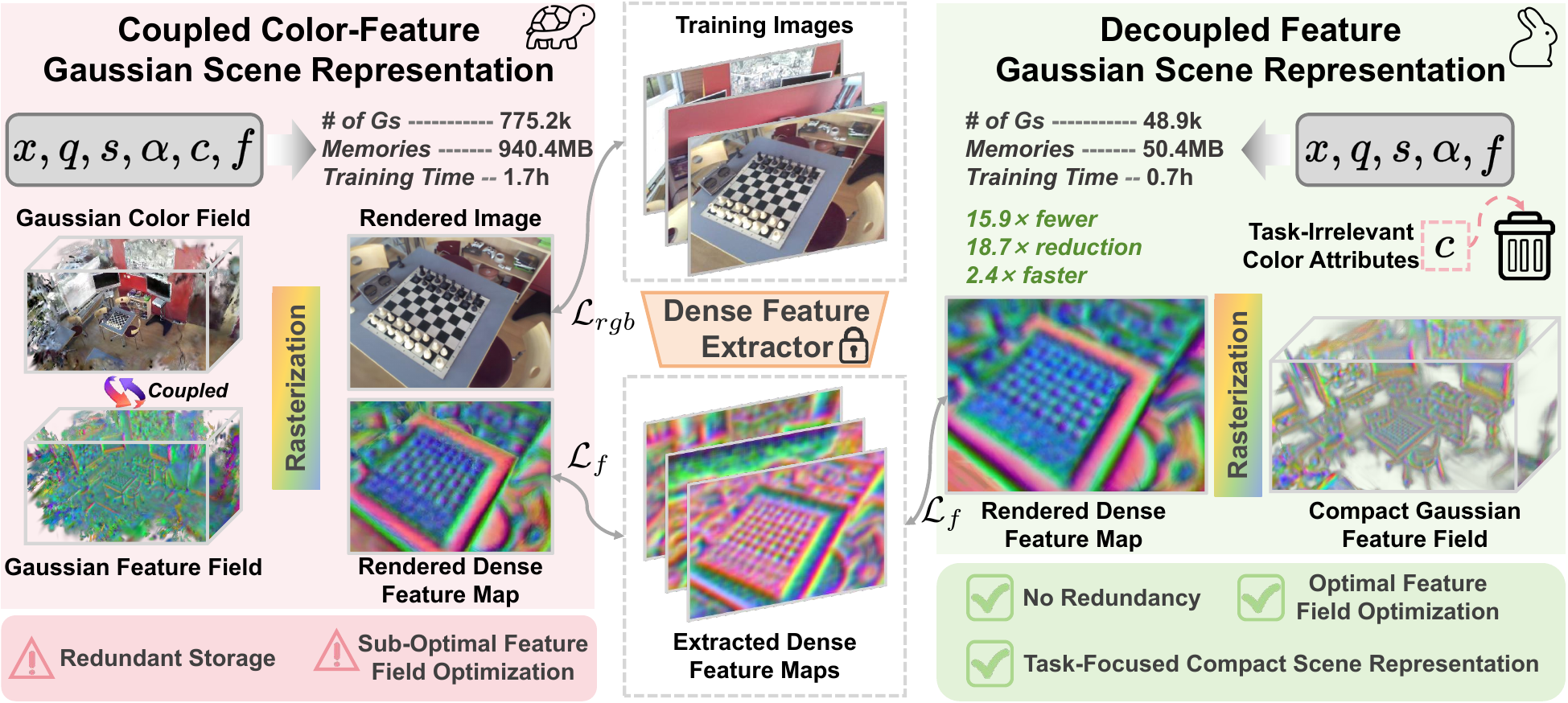}
	\vspace{-0.2in}
	\caption{Coupled Color–Feature vs. Decoupled Feature Gaussian Fields.}
	\label{fig:decoupled_feature_field}
\end{figure}

Despite its encouraging localization accuracy, STDLoc is intrinsically haunted by two critical efficiency bottlenecks. First, STDLoc adopts the standard Feature 3DGS, which employs a coupled representation where color and feature attributes share the same Gaussian primitives. However, (1) the color field is functionally irrelevant for visual localization, as pose estimation relies on geometric constraints provided by 2D–3D matches rather than photometric appearance; and (2) a pronounced frequency mismatch exists: high-frequency photometric details entail excessively dense Gaussian primitives, compelling the storage of redundant high-dimensional embeddings for the inherently low-frequency feature field. This coupling leads to prohibitive memory overhead and sub-optimal feature field optimization, as shown in Fig.~\ref{fig:decoupled_feature_field}. Second, at the pose refinement stage, tens of thousands spatially correlated dense matches are fed into the PnP solver. Practically, most of them provide highly saturated geometric constraints, yielding marginal accuracy gains while imposing a substantial computational burden on the robust estimator.

\begin{figure}[t]
	\centering
	\includegraphics[width=1\linewidth]{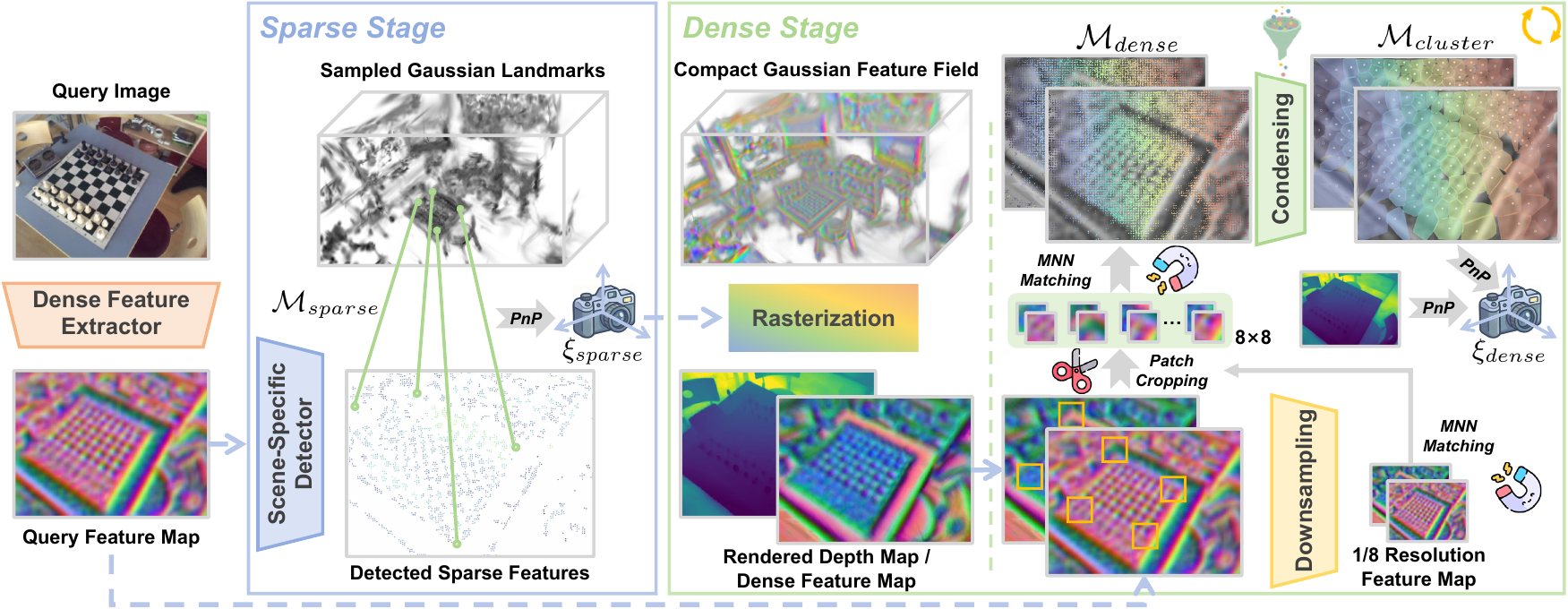}
	\vspace{-0.2in}
	\caption{The framework of our LiteLoc. Zoom in for better visualization.}
	\label{fig:pipeline}
\end{figure}

To address these limitations, we present LiteLoc, an innovative and efficient sparse-to-dense localizer with compact 3DGS scene representation and accelerated dense refinement. First, we introduce a color-free decoupled feature field tailored for localization. By removing all color-related attributes, such as spherical harmonics (SH) coefficients, from 3D Gaussians, LiteLoc allows the Gaussian density to align the intrinsic low-frequency structure of the feature field rather than the high-frequency color field. This avoids binding large feature embeddings to photometrically driven Gaussians and reduces storage by $\approx 94\%$ without sacrificing localization-relevant information, as shown in Fig.~\ref{fig:decoupled_feature_field}. Second, we deploy a dense match condensing mechanism. Instead of solving PnP on all spatially correlated dense matches, LiteLoc performs joint geometric-feature clustering to retain only an $\approx 5\%$ subset of representative matches that preserve geometric diversity, as shown in Fig.~\ref{fig:pipeline}. This removes saturated constraints and accelerates robust pose estimation by $\approx19\times$ while maintaining accuracy. Together, these designs empower LiteLoc with significantly lower memory footprint, faster inference, and improved localization accuracy over STDLoc across challenging benchmarks like 7-Scenes.

{\bf Methodology:} 1) \emph{Pipeline Overview}: As shown in Fig.~\ref{fig:pipeline}, LiteLoc inherits the sparse-to-dense hierarchical localization architecture of STDLoc~\cite{huang2025sparse}. The pipeline consists of a mapping phase, where a 3D Gaussian scene representation ${\mathcal{G}}$ is constructed, followed by a two-stage localization phase. In the sparse stage, to ensure reliable matching against Gaussian primitives, the matching-oriented sampling strategy~\cite{huang2025sparse} is first employed to filter out ambiguous primitives and select a subset of discriminative landmarks $\tilde{\mathcal{G}}$ that are multi-view consistent and evenly distributed. Then, given a query image $I_q$, a generic local feature descriptor, such as SuperPoint~\cite{detone2018superpoint}, is utilized to generate a dense query feature map, followed by a scene-specific detector~\cite{huang2025sparse}—trained in a self-supervised manner by projecting sampled landmarks onto training views—to detect sparse local features. These features are matched against the landmarks $\tilde{\mathcal{G}}$ via cosine similarity to establish sparse 2D-3D correspondences $\mathcal{M}_{sparse}$, which are then used to solve the coarse pose $\xi_{sparse}$ via a PnP solver. In the subsequent dense stage, utilizing $\xi_{sparse}$, we render both a dense feature map $\hat{F}_s$ and a depth map $\hat{D}$ from the Gaussian feature field. The rendered features $\hat{F}_s$ are aligned with the query's feature map through coarse-to-fine matching to establish dense 2D-2D correspondences $\mathcal{M}_{dense}$, which are lifted to 2D-3D using $\hat{D}$ to solve for the final refined pose $\xi_{dense}$. The dense stage allows for iterative pose refinement to achieve higher accuracy. While following this paradigm, LiteLoc fundamentally reconstructs the scene representation and accelerates the dense solver as follows.

2) \emph{Color-free Decoupled Feature Field}: As shown in Fig.~\ref{fig:decoupled_feature_field}, STDLoc adopts a coupled Gaussian representation~\cite{zhou2024feature}, parameterizing each primitive as $\Theta_i = \{x_i, q_i, s_i, \alpha_i, c_i, f_i\}$. Here, $x_i, q_i, s_i, \alpha_i$ represent position, rotation, scaling, and opacity, respectively; $c_i$ denotes the SH coefficients for modeling view-dependent color; and $f_i$ is the feature. This representation entangles geometric optimization with both high-frequency photometric reconstruction and low-frequency feature field consistency. Yet, we observe that reconstructing photometric details via $c_i$ necessitates excessive primitives, forcing the storage of redundant high-dimensional embeddings for the inherently low-frequency feature field. To address this, LiteLoc introduces a color-free decoupled feature field, as shown in Fig.~\ref{fig:decoupled_feature_field}. We construct a compact Gaussian field parameterized solely by $\Theta'_i = \{x_i, q_i, s_i, \alpha_i, f_i\}$, explicitly discarding $c_i$. To ensure stability and robustness in the feature field training and rendering, the feature map $\hat{F}$ is rendered via $L_{2}$-normalized alpha blending:
\begin{equation}
	\hat{F} = \text{norm}\left( \sum_{i \in \mathcal{N}} \text{norm}(f_i) \alpha_i \prod_{j=1}^{i-1}(1-\alpha_j) \right),
	\label{eq:rendering}
\end{equation}
where $\text{norm}(\cdot)$ denotes $L_{2}$ normalization and $\mathcal{N}$ is the set of Gaussians associated with a pixel, sorted in front-to-back order. We supervise the field with SuperPoint feature map $F_{gt}$ using $\mathcal{L}_{f} = \|F_{gt} - \hat{F}\|_1$. This decoupling allows the Gaussian density to adapt strictly to the feature structure rather than appearance, reducing storage overhead by $\approx 90\%$. Also, liberating the geometry from high-frequency color constraints facilitates better optimization convergence for the feature field, leading to a more robust scene representation.

3) \emph{Dense Match Condensing}: In the dense stage, STDLoc performs a coarse-to-fine matching process between the rendered feature map $\hat{F}_s$ and the query feature map. Matches are first established at a low resolution (\emph{i.e.}, $1/8$ scale) by computing a dual-softmax probability matrix on the correlation of coarse features, followed by mutual nearest neighbor (MNN) selection. These coarse matches are then refined to pixel-level precision by correlating local windows (\emph{i.e.}, $8\times8$) in the high-resolution feature maps. This process typically yields a massive set of dense correspondences $\mathcal{M}_{dense}$ (often $\vert\mathcal{M}\vert > 10^4$), as shown in Fig.~\ref{fig:pipeline}. Feeding this entire set into a PnP solver causes high latency due to saturated geometric constraints. Thus, we propose a condensing strategy to distill these raw matches into a minimal set of representative ones. Specifically, we embed the dense correspondences into a joint 4D geometric space formed by the 2D coordinates of the query and rendered views. We partition this space into $K$ distinct regions $\{\mathcal{C}_k\}_{k=1}^K$ via K-means clustering. The correspondence nearest to the centroid of each cluster is selected as the representative proxy $\mathcal{M}_{cluster}^{k}=(u_k, \bar{u}_k)$, effectively pruning spatial redundancy while preserving geometric distribution. The proxy point $\bar{u}_k$ in the rendered view is back-projected to a 3D point $P_k$ using the rendered depth $\hat{D}$. Consequently, the pose estimation is formulated as solving the optimal rigid transformation $\xi^* = \{R^*, t^*\}$ that minimizes the robust reprojection error over these proxies:
\begin{equation}
	R^*, t^* = \operatorname*{arg\,min}_{R, t} \sum_{k=1}^{K} \rho \left( \| u_k - \pi(R P_k + t) \|_2 \right),
	\label{eq:pose}
\end{equation}
where $\pi(\cdot)$ denotes the perspective projection function and $\rho(\cdot)$ is the robust kernel. We set $K = 1024 \approx 0.05\vert\mathcal{M}\vert$ as a cost-effective default, which balances geometric coverage and computational efficiency, achieving nearly a $19\times$ speedup in robust estimation while maintaining high-precision localization. Empirically, the condensing strategy remains robust to varying $K$ across diverse scenes.

\begin{table}[t]
	\centering
	\scriptsize
	\setlength{\tabcolsep}{2.5pt}
	\renewcommand{\arraystretch}{1.20}
	\caption{Localization results on 7-Scenes. The \rkone{\text{best}}, \rktwo{\text{second-best}} and \rkthree{\text{third-best}} results are highlighted.}
	\label{tab:7scenes}
	\vspace{-0.05in}
	\resizebox{\columnwidth}{!}{
		\begin{tabular}{l l c c c c c c c c}
			\toprule
			& Method & Chess & Fire & Heads & Office & Pumpkin & RedKitchen & Stairs & Avg.$\downarrow$ [cm/°] \\
			\midrule
			\multirow{2}{*}{\rotatebox{90}{FM}}
			& AS (SIFT)       
			& 3/0.87    
			& 2/1.01    
			& 1/0.82    
			& 4/1.15    
			& 7/1.69    
			& 5/1.72    
			& 4/1.01    
			& 3.71/1.18 \\
			
			& HLoc (SP+SG)    
			& 2.39/0.84 
			& 2.29/0.91 
			& 1.13/0.77 
			& 3.14/0.92 
			& 4.92/1.30 
			& 4.22/1.39 
			& 5.05/1.41 
			& 3.31/1.08 \\
			
			\hdashline
			
			\multirow{3}{*}{\rotatebox{90}{SCR}}
			& DSAC* & \rkthree{0.50}/\rkthree{0.17} 
			& \rkthree{0.78}/\rkthree{0.29} 
			& \rkthree{0.50}/0.34 
			& 1.19/0.35 
			& 1.19/0.29 
			& \rkthree{0.72}/\rkthree{0.21} 
			& \rkthree{2.65}/\rkthree{0.78} 
			& \rkthree{1.07}/0.35 \\ 
			
			& ACE             
			& 0.55/0.18 
			& 0.83/0.33 
			& 0.53/\rkthree{0.33} 
			& \rkthree{1.05}/\rkthree{0.29} 
			& \rkthree{1.06}/\rkthree{0.22} 
			& 0.77/\rkthree{0.21}  
			& 2.89/0.81 
			& 1.10/\rkthree{0.34} \\ 
			
			& NeuMap          
			& 2/0.81 
			& 3/1.11 
			& 2/1.17 
			& 3/0.98 
			& 4/1.11 
			& 4/1.33 
			& 4/1.12 
			& 3.14/0.95 \\
			
			\hdashline
			
			\multirow{6}{*}{\rotatebox{90}{NeRF/GS}}
			& DFNet+NeFeS$_{50}$ 
			& 2/0.57 
			& 2/0.74 
			& 2/1.28 
			& 2/0.56 
			& 2/0.55 
			& 2/0.57 
			& 5/1.28 
			& 2.43/0.79 \\
			
			& CROSSFIRE       
			& 1/0.40 
			& 5/1.90 
			& 3/2.30 
			& 5/1.60 
			& 3/0.80 
			& 2/0.80 
			& 12/1.90 
			& 4.43/1.38 \\
			
			& PNeRFLoc        
			& 2/0.80 
			& 2/0.88 
			& 1/0.83 
			& 3/1.05 
			& 6/1.51 
			& 5/1.54 
			& 32/5.73 
			& 7.29/1.76 \\
			
			& NeRFMatch       
			& 0.95/0.30 
			& 1.11/0.41 
			& 1.34/0.92 
			& 3.09/0.87 
			& 2.21/0.60 
			& 1.03/0.28 
			& 9.26/1.74 
			& 2.71/0.73 \\
			
			& STDLoc          
			& \rktwo{0.46}/\rktwo{0.15} 
			& \rktwo{0.57}/\rktwo{0.24} 
			& \rktwo{0.45}/\rktwo{0.26} 
			& \rktwo{0.86}/\rktwo{0.24} 
			& \rktwo{0.93}/\rktwo{0.21} 
			& \rktwo{0.63}/\rktwo{0.19} 
			& \rkone{1.42}/\rktwo{0.41} 
			& \rktwo{0.76}/\rktwo{0.24} \\ 
			
			& \textbf{LiteLoc (Ours)}   
			& \rkone{0.45}/\rkone{0.13} 
			& \rkone{0.49}/\rkone{0.20} 
			& \rkone{0.40}/\rkone{0.25} 
			& \rkone{0.75}/\rkone{0.20} 
			& \rkone{0.88}/\rkone{0.20} 
			& \rkone{0.62}/\rkone{0.14} 
			& \rktwo{1.50}/\rkone{0.40} 
			& \rkone{0.73}/\rkone{0.22} \\ 
			
			\midrule
			
			& LiteLoc (w/o Cond.) 
			& 0.46/0.14
			& 0.47/0.20
			& 0.40/0.25
			& 0.74/0.20
			& 0.88/0.20
			& 0.62/0.14
			& 1.49/0.39
			& 0.72/0.22 \\
			
			\bottomrule
		\end{tabular}
	}
\end{table}

\begin{table}[t]
	\centering
	\scriptsize
	\setlength{\tabcolsep}{3pt}
	\renewcommand{\arraystretch}{1.2}
	\caption{Localization results on Cambridge Landmarks.}
	\label{tab:cambridge}
	\vspace{-0.05in}
	\resizebox{\columnwidth}{!}{%
		\begin{tabular}{l l c c c c c c}
			\toprule
			& Method & Court & King's & Hospital & Shop & St. Mary's & Avg.$\downarrow$ [cm/°] \\
			\midrule
			\multirow{2}{*}{\rotatebox{90}{FM}}
			& AS (SIFT)
			& 24/0.13
			& \rkthree{13}/0.22
			& 20/0.36
			& \rktwo{4}/\rkthree{0.21}
			& 8/\rkthree{0.25}
			& 13.8/0.23 \\
			
			& HLoc (SP+SG)
			& 17.7/0.11
			& \rkone{11}/0.20
			& \rkthree{15.1}/\rkthree{0.31}
			& \rkthree{4.2}/\rktwo{0.20}
			& \rktwo{7}/\rktwo{0.22}
			& \rkthree{11.0}/\rkthree{0.21} \\
			
			\hdashline
			
			\multirow{3}{*}{\rotatebox{90}{SCR}}
			& DSAC*
			& 33/0.21
			& 17.9/0.31
			& 21.1/0.40
			& 5.2/0.24
			& 15.4/0.51
			& 18.5/0.33 \\
			
			& ACE (Poker)
			& 27.9/0.14
			& 18.6/0.33
			& 25.7/0.51
			& 5.1/0.26
			& 9.5/0.33
			& 17.4/0.31 \\
			
			& NeuMap
			& \rkone{6}/0.10
			& 14/\rkthree{0.19}
			& 19/0.36
			& 6/0.25
			& 17/0.53
			& 12.4/0.29 \\
			
			\hdashline
			
			\multirow{7}{*}{\rotatebox{90}{NeRF/GS}}
			& DFNet+NeFeS$_{50}$
			& -
			& 37/0.54
			& 52/0.88
			& 15/0.53
			& 37/1.14
			& 35.3/0.77 \\
			
			& CROSSFIRE
			& -
			& 47/0.70
			& 43/0.70
			& 20/1.20
			& 39/1.40
			& 37.3/1.00 \\
			
			& PNeRFLoc
			& 81/0.25
			& 24/0.29
			& 28/0.37
			& 6/0.27
			& 40/0.55
			& 35.8/0.35 \\
			
			& NeRFMatch
			& 19.6/\rkthree{0.09}
			& \rktwo{12.5}/0.23
			& 20.9/0.38
			& 8.4/0.40
			& 10.9/0.35
			& 14.5/0.29 \\
			
			& STDLoc
			& \rkthree{15.7}/\rktwo{0.06}
			& 15.0/\rktwo{0.17}
			& \rktwo{11.9}/\rktwo{0.21}
			& \rkone{3.0}/\rkone{0.13}
			& \rkone{4.7}/\rkone{0.14}
			& \rktwo{10.1}/\rkone{0.14} \\
			
			& \textbf{LiteLoc (Ours)}
			& \rktwo{7.92}/\rkone{0.05}
			& 15.41/\rkone{0.16}
			& \rkone{9.02}/\rkone{0.19}
			& 4.66/0.26
			& \rkthree{7.67}/0.31
			& \rkone{8.90}/\rktwo{0.19} \\
			
			\midrule
			
			& LiteLoc (w/o Cond.) 
			& 7.96/0.05
			& 15.90/0.18
			& 9.14/0.21
			& 4.60/0.25
			& 7.60/0.31
			& 9.04/0.20 \\
			
			\bottomrule
		\end{tabular}%
	}
\end{table}

\begin{table}[t]
	\centering
	\scriptsize
	\setlength{\tabcolsep}{1.5pt} 
	\renewcommand{\arraystretch}{1.25} 
	
	\caption{Ablation study and efficiency analysis on 7-Scenes.}
	\label{tab:ablation}
	\vspace{-0.1in}
	
	\resizebox{\columnwidth}{!}{%
		\begin{tabular}{l cc cc cc ccccc}
			\toprule
			\multirow{2.5}{*}{\textbf{Method}} & 
			\multicolumn{2}{c}{\textbf{Design}} & 
			\multicolumn{2}{c}{\textbf{Accuracy} (\%)\thinspace$\uparrow$} & 
			\multicolumn{2}{c}{\textbf{Storage}\thinspace$\downarrow$} & 
			\multicolumn{5}{c}{\textbf{Time (ms)}\thinspace$\downarrow$} \\
			
			\cmidrule(lr){2-3} \cmidrule(lr){4-5} \cmidrule(lr){6-7} \cmidrule(lr){8-12}
			
			& Decoup. & Cond. 
			& 5cm~5$^\circ$ & 2cm~2$^\circ$ 
			& \#Gs (k) & Mem (MB) 
			& Rast. & Clus. & PnP & Total${_1}$ & Total${_4}$\\
			\midrule
			
			STDLoc
			& \xmark & \xmark
			& 99.1 & 90.9
			& \cellN{759.4} & \cellN{929.5}
			& \cellN{16.4} & -   & \cellN{96.4} & \cellN{230.2} & \cellN{613.1} \\
			
			STDLoc w/ Cond.
			& \xmark & \cmark
			& 99.1 & 90.7
			& \cellN{759.4} & \cellN{929.5}
			& \cellN{16.4} & 8.8 & \cellN{5.2} & \cellN{\textbf{147.5}} & \cellN{\textbf{283.3}} \\
			
			LiteLoc w/o Cond.
			& \cmark & \xmark
			& \textbf{99.6} & \textbf{92.4}
			& \cellN{\textbf{57.0}} & \cellN{\textbf{58.8}}
			& \cellN{\textbf{9.3}} & - & \cellN{96.5} & \cellN{257.8} & \cellN{619.8} \\
			
			\rowcolor{Gray}
			\textbf{LiteLoc (Ours)}
			& \cmark & \cmark
			& \textbf{99.6} & 92.2
			& \cellD{\textbf{57.0}}{92} & \cellD{\textbf{58.8}}{94}
			& \cellD{\textbf{9.3}}{43} & 8.8 & \cellD{\textbf{5.1}}{95}
			& \cellD{174.5}{24} & \cellD{287.3}{53} \\
			
			\bottomrule
		\end{tabular}%
	}
\end{table}

{\bf Experiments:} 1) \emph{Training and Localization Details}: LiteLoc is trained per-scene following STDLoc~\cite{huang2025sparse}.
The decoupled feature field is optimized for 50000 steps with learning rate 0.001; the remaining training settings follow STDLoc. At inference, to adapt the matching-oriented sampling strategy of STDLoc to LiteLoc's compact Gaussian field, we reduce the number of nearest neighbors to 6, ultimately sampling 16384 anchors to maintain spatial coverage. Dense pose refinement is iterated 4/1 times on 7-Scenes/Cambridge Landmarks. For dense match condensing, K-means clustering is performed using FAISS~\cite{douze2025faiss} on CPU, with at most 5 iterations. All experiments are conducted on one NVIDIA RTX~4090 GPU.

2) \emph{Benchmarks and Metrics}:
We evaluate LiteLoc on 7-Scenes~\cite{shotton2013scene} and Cambridge Landmarks~\cite{kendall2015posenet} datasets, and report the \emph{median} translation and rotation errors (cm/$^\circ$) per scene.

3) \emph{Baselines}: We compare against FM-based methods (\emph{e.g.}, AS~\cite{sattler2016efficient}, HLoc~\cite{sarlin2019coarse}), SCR-based methods (\emph{e.g.}, DSAC*~\cite{brachmann2021visual}, ACE~\cite{brachmann2023accelerated }, NeuMap~\cite{tang2023neumap}), and recent NeRF/GS-based localizers~\cite{chen2024neural,moreau2023crossfire,zhao2024pnerfloc,zhou2024nerfect,huang2025sparse}.

4) \emph{Indoor Localization on 7-Scenes}: Table~\ref{tab:7scenes} shows that LiteLoc achieves the best overall accuracy among NeRF/GS-based methods on 7-Scenes, reducing the average error from 0.76/0.24 (STDLoc) to 0.73/0.22. Notably, LiteLoc yields consistent gains across most scenes (\emph{e.g.}, Chess, Fire, Heads, Office, Pumpkin, RedKitchen), while maintaining comparable performance on Stairs. These results indicate that removing photometric attributes and compacting the feature field does \emph{not} compromise pose accuracy; instead, it can even improve robustness by producing a cleaner, task-focused scene representation.

5) \emph{Outdoor Localization on Cambridge Landmarks}: As shown in Table~\ref{tab:cambridge}, LiteLoc also performs competitively on large-scale outdoor localization. Compared with STDLoc (Avg.\ 10.1/0.14), LiteLoc reduces the average translation error to 8.9cm, while maintaining a comparable rotation error (0.19$^\circ$). LiteLoc achieves the best translation accuracy on Court and Hospital, and remains strong on King's and St.\ Mary's, demonstrating that the proposed efficiency-oriented redesign remains effective on outdoor scenes.

6) \emph{Ablations and Efficiency Analysis}:
Table~\ref{tab:ablation} validates the contribution of each design choice and quantifies the resulting efficiency gains.
Since LiteLoc mainly targets the dense refinement bottleneck, we explicitly decompose the dense-stage runtime into rasterization, clustering, and PnP.
We also report the \emph{total localization runtime}, including the sparse stage and either a \emph{single} dense refinement iteration (Total$_1$) or \emph{four} dense refinement iterations (Total$_4$), matching the evaluation protocol on 7-Scenes.
Notably, the overall inference cost scales with the number of dense refinement iterations.

\emph{(i) Decoupled Feature Field.}
Enabling decoupling alone dramatically compresses the scene representation:
the number of Gaussians is reduced from 759.4k to 57.0k and storage from 929.5MB to 58.8MB,
while simultaneously improving accuracy (from 99.1/90.9 to 99.6/92.4 under 5cm~5$^\circ$/2cm~2$^\circ$).
This compact representation also reduces rasterization time from 16.4ms to 9.3ms.
However, without condensing, the dense PnP stage remains the dominant bottleneck, resulting in limited end-to-end speedup.

\emph{(ii) Dense Match Condensing.}
Applying dense match condensing alone directly targets the dominant bottleneck in STDLoc.
The dense PnP time is reduced from 96.4ms to 5.2ms, with a modest clustering overhead of 8.8ms.
As a result, the total runtime per iteration (Total$_1$) decreases from 230.2ms to 147.5ms,
and the full four-iteration runtime (Total$_4$) drops from 613.1ms to 283.3ms,
while preserving localization accuracy (99.1/90.7).

\emph{(iii) Full LiteLoc.}
Combining both designs, LiteLoc achieves high accuracy (99.6/92.2) with only 58.8MB storage (94\% reduction), while reducing the total runtime to 174.5ms for a single dense iteration and 287.3ms for four iterations. The dense solver is accelerated by nearly $19\times$ (PnP: 96.4$\rightarrow$5.1\,ms), leading to a 53\% reduction in full inference time (Total$_4$).
Overall, LiteLoc strikes a satisfying balance between efficiency and accuracy on practical visual localization.

{\bf Conclusion:}
This letter introduces LiteLoc, an efficient sparse-to-dense visual localization framework powered by 3DGS. By leveraging a color-free decoupled feature field and a dense match condensing strategy, LiteLoc substantially reduces storage overhead and inference latency without compromising accuracy. Extensive experiments demonstrate the effectiveness of LiteLoc for practical localization.



\enlargethispage{2\baselineskip}

\end{document}